\documentclass[conference]{IEEEtran}
\IEEEoverridecommandlockouts
\usepackage{cite}
\usepackage{amsmath,amssymb,amsfonts}
\usepackage{algorithmic}
\usepackage{graphicx}
\usepackage{textcomp}
\usepackage{xcolor}

\usepackage{url}
\usepackage[hidelinks]{hyperref}
\usepackage[utf8]{inputenc}
\usepackage[small]{caption}
\usepackage{booktabs}
\usepackage{algorithm}
\usepackage{algorithmic}
\usepackage{amsmath}
\usepackage{bm}
\newcommand{\m}[1]{\bm{#1}}

\usepackage{algorithm}
\usepackage{algorithmic}
\usepackage{booktabs}

\usepackage{siunitx}

\usepackage{amssymb, amsmath}
\usepackage{graphicx}
\usepackage{hyperref}
\usepackage{cleveref}
\usepackage{amsthm}

\newcommand{\doo}{\textrm{do}}

\def\BibTeX{{\rm B\kern-.05em{\sc i\kern-.025em b}\kern-.08em
    T\kern-.1667em\lower.7ex\hbox{E}\kern-.125emX}}
\begin{document}

\title{Active learning of causal probability trees}

\author{\IEEEauthorblockN{Tue Herlau}
	\IEEEauthorblockA{\textit{DTU Compute} \\
		\textit{Technical University of Denmark}\\
		2800 Lyngby, Denmark \\
		tuhe@dtu.dk}	
}

\maketitle

\begin{abstract}
	The past two decades have seen a growing interest in combining causal information, commonly represented using causal graphs, with machine learning models. Probability trees provide a simple yet powerful alternative representation of causal information. They enable both computation of intervention and counterfactuals, and are strictly more general, since they allow context-dependent causal dependencies. Here we present a Bayesian method for learning probability trees from a combination of interventional and observational data. The method quantifies the expected information gain from an intervention, and selects the interventions with the largest gain. We demonstrate the efficiency of the method on simulated and real data. An effective method for learning probability trees on a limited interventional budget will greatly expand their applicability.
\end{abstract}

\begin{IEEEkeywords}
	Causal learning; Bayesian methods; active learning
\end{IEEEkeywords}

\section{Introduction}
The past two decades have seen an increasing use of causal reasoning within fairness\cite{ijcai2019-199,chiappa2019path}, AI safety~\cite{everitt2021agent}, medicine~\cite{richens2020improving,prosperi2020causal}, and reinforcement learning~\cite{bareinboim2015bandits,dasgupta2019causal,de2019causal,yue2020interventional}. This is thanks to its ability to model and predict relationships that are not statistical, such as the result of interventions and counterfactual queries, and the increasing understanding that these relationships are important in AI and machine learning~\cite{pearl2018book}. As a consequence, \emph{causal induction}, that is learning causal relationships in the first place, can be expected to play an increasingly important role in machine learning~\cite{scholkopf2019causality}.

\begin{figure}[t!]
	\centering
	\includegraphics[width=.95\linewidth]{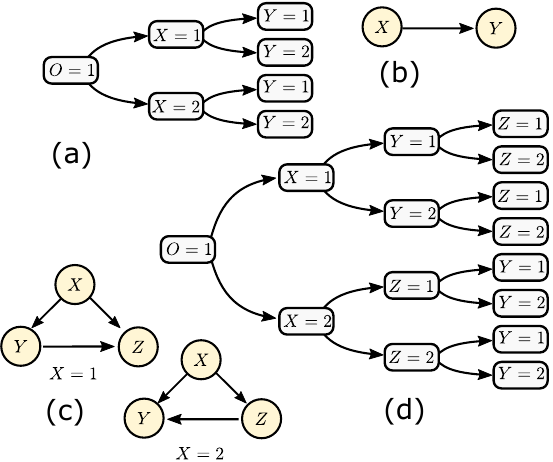}
	\caption{(a) Example of a probability tree for two binary variables $X$, $Y$, each path from the root $O$ to a leaf denotes an assignment of the variables and (b) the corresponding CBN. (c--d) Probability trees can model context-dependent causal dependency, in this case the value of $X$ determines the causal orientation of $Y$ and $Z$. }\label{fig2svg2}
\end{figure}

Causal relationships are traditionally described using structural causal models (SCMs)~\cite{pearl2000models} or causal Bayesian networks (CBNs)~\cite{spirtes2000causation}, both of which represent the statistical independence properties implied by the causal model as a graph. Although versatile, both SCMs and CBNs are limited by the assumption that causal relationships between the observed variables must follow a partial order~\cite{pmlr-v6-dawid10a}.
This work considers a more general representation of causal relationships, namely a discrete probability tree~\cite{shafer1996art} (see \cref{fig2svg2}). Probability trees are able to express context-specific dependencies (that is, situations where the causal order depends on the value of variables in the causal model, see \cref{fig2svg2} (c))). Recent work has shown how interventions and counter-factual queries can be computed from a probability tree~\cite{genewein2020algorithms}, but does not provide a method for learning probability trees in the first place.

In this paper, we consider the problem of efficiently learning causal relationships, as represented by probability trees, from a combination of observational and interventional data. 
Specifically, we consider an active-learning setting where an agent must decide which one of the available interventions should be attempted in each step, so as to learn which probability tree represents the generative process of the data. Although this problem has previously been studied for both SCMs and CBNs, this is, to our knowledge, the first work which considers the problem for probability trees. 

Our approach combines two ideas: First, we make use of the ability of probability trees to represent context-dependent relationships by representing the different causal hypothesis as sub-trees in a single large probability tree, thereby reducing the problem of causal induction to a simple inference problem in this larger probability tree, which can be solved using Bayes theorem; second, by combining the causal hypotheses in a single model, we can predict the information gain associated with each intervention in advance, and thus select the intervention which has the highest gain, thereby leading to a natural active-learning method for causal induction on probability trees\footnote{Code to reproduce all plots in this paper can be found at \url{http://github.com/anonymized_for_review}}.

\paragraph{Related work}
Information theory has previously been used to quantify the causal effect between variables~\cite{wieczorek2019information}, or to specify circumstances where the causal orientation of categorical variables can be determined from observational data~\cite{compton2021entropic}. Information geometry was used to infer causal orientation from observational data, using assumptions on generative mechanisms~\cite{janzing2012information}, however both settings are different from the Bayesian learning problem considered here.
Most relevant to our work is the active-learning method discussed in~\cite{tong2001active}, where interventions are selected by their expected reduction of a certain cost-function, defined using the entropy of the edge-distribution. The method is used as a point of comparison in \cref{sec:simulations}. Context-dependent relationships occur naturally in a variety of places, such as probabilistic programming~\cite{vajda2014probabilistic}, or as an additional node type in factor graphs~\cite{winn2012causality}.

\section{Methods} 
A probability tree describes the generative process of the data and is easiest defined recursively. We will be concerned with $m$ discrete stochastic variables $\m X = (X_1, \dots, X_m)$, so that $X_\ell \in \mathcal V_\ell$ is the range of values of $X_\ell$. A probability tree describes the generative process of $\m X$ as a path through the tree, starting at the root, where at each node of the tree one or more variables in $\m X$ is assigned its value. Suppose $n \in \mathcal T$ is a node in the tree. We represent $n$ as a tuple $n = (u, \mathcal S, \theta^{(n)})$ where $u$ is a unique integer identifying the node, $\mathcal S$ is a list of statements specifying which variables are assigned at $n$ (in all our examples, just a single statement of the form $X_\ell = x_\ell$) and $\theta^{(n)}_1,\dots, \theta^{(n)}_{|\textrm{ch}(n)|} \in [0,1]$ is the transition probability from $n$ to each of its children $\textrm{ch}(n)$. For completeness, the root will be associated with the dummy variable $O=1$, which can only take a single value. 

A \emph{total realization} is a path $\tau = (n_1, \dots, n_l)$ from the root to a leaf in the tree. We assume that the tree is constructed such that the statements associated with the nodes on any path form a partition of all variables $X_1, \dots, X_m$; that is, when traversing from the root to the leaf, all variables are assigned a value once. The probability of a realization is simply the product of the probabilities $\theta^{(n)}_i$ of each node encountered along the path. Examples of Bayesian networks and their corresponding probability trees are given in \cref{fig2svg2} (a--b).

An \emph{event}, such as $X_k = x_k$ or $\{X_1 = 3, X_5 = 7\}$, is identified with the set of all total realizations which traverses nodes with statements $X_k=x_k$. The probability of an event is the sum of probability of the total realizations, thereby defining the joint distribution $P(X_1, \dots, X_m)$.

\subsection{Intervention}\label{sec2.1}
Interventions in probability trees are defined following~\cite{genewein2020algorithms}: An intervention $X_\ell = x_\ell$ in a probability tree simply means that during the generative process, at the node where the value $X_\ell$ is decided, the path is forced to choose the branch corresponding to $X_\ell = x_\ell$ with probability 1. Formally, an intervention on $\mathcal T$ can be defined as a new tree $\mathcal T'$ identical to $T$, except for all nodes $n = (v,\mathcal{S}, \theta^{(n)})$ which have a child in which $X_\ell$ is assigned a value, we change $\theta^{(n)}$ so if the child has an assignment $X_\ell \neq x_\ell$ it is selected with probability 0, and if the child has a statement of the form $X_\ell = x_\ell$ it is selected with probability 1. 

\subsection{Causal hypothesis}
\begin{figure}[t!]
	\centering
	\includegraphics[width=.95\linewidth]{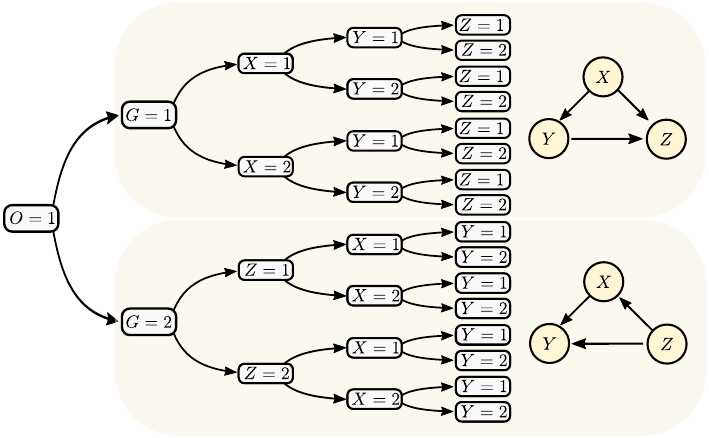}
	\caption{A single probability tree with a context-dependent variable $G$ can represent the $S=2$ causal hypothesis for the variables $X,Y,Z$. Depending on the value of $G$, the causal ordering of $X,Y,Z$ will differ, as illustrated by the two CBNs. }\label{fig3svg}
\end{figure}

An appealing feature of probability trees is that selecting between different causal hypotheses, i.e. the causal orderings of the variables $X_1,\dots,X_m$, can be treated as a context-dependent causal problem. Suppose $\mathcal T_1,\dots, \mathcal T_S$ are trees of the form described earlier, which denote different causal hypotheses for the variables $X_1,\dots,X_m$ and we let $G=1,\dots,S$ be a stochastic variable denoting each hypothesis. We can then represent the problem using a new tree, where the root has $S$ children, each of the form $n_k = (k, G = k, \theta^{(n_k)})$, and corresponding to one of the trees $\mathcal T_k$. An illustration is given in \cref{fig3svg} using $S=2$ causal hypothesis over three variables $X,Y,Z$, where the plates indicate the two different causal hypotheses shown as CBNs. A total realization in the new tree $\mathcal{T}$ then corresponds both to selecting a causal orientation, $G=k$, and a realization of the variables $X_1,\dots,X_m$ according to $\mathcal T_k$. 

\subsection{Bayesian learning}
We will be concerned with the case where interventions occur on single variables, i.e. are of the form $J = \{X_\ell = x\}$, and the empty intervention $J = \emptyset$ will correspond to passively observing the system. The available data $D$ is therefore a sequence of interventions and the realized values of the $m$ variables $D = (\mathbf X^{(1)}, J^{(1)}), \dots, (\mathbf X^{(N)}, J^{(N)})$. 

Assuming a uniform prior over the $S$ causal hypothesis, $\theta^{(n_O)} = P(G=k) = \frac{1}{S}$, the probability of a given causal order is immediately available using Bayes theorem
\begin{align}
	P(G=k | D, \m \theta) = \frac{ P(D | G=k, \m \theta) }{ \sum_{k' = 1}^{ S } P(D | G=k', \m \theta) } \label{eq:1a}
\end{align}
where $\bm \theta = (\theta^{(n)})_{n \in \mathcal{T}_k}$ are the split probabilities. To compute the marginal likelihoods, note that each observation $\mathbf X^{(i)} = (X_1^{(i)}, \dots, X_m^{(i)})$ in the data set corresponds to a total realization of each of the trees $\mathcal T_k$, and the probability can therefore be obtained by simply using the generative procedure outlined earlier. Specifically, for a node $n \in \mathcal T_k$, we let $N_{j|n}$ be the total number of realizations passing through the node $n$ and selecting child $j \in \textrm{ch}(n)$, but \emph{not} counting those of the total realizations, where the choice $j$ was forced by the intervention (see  \cref{sec2.1}). The probability of the data conditional on tree $\mathcal T_k$ is therefore:
\begin{align}
	P(D | G=k, \m \theta ) =  \prod_{n\in \mathcal{T}_k } \left[ \prod_{j \in \textrm{ch}(n)} 
	\left( \theta^{(n)}_j \right)^{ 
		N_{j|n}  } \right].\label{eq:5}
\end{align}
Assuming each $\theta^{(n)} \sim \textrm{Dir}(\alpha^{(n)} \m 1_{|\textrm{ch}(n)|\times 1| } )$ has a Dirichlet prior with concentration parameter $\alpha^{(n)}$, the marginal likelihood is obtained by integrating \cref{eq:5}: 
\begin{align}
	P(D | G=k ) & = \int \prod_{n\in \mathcal{T}_k } d\theta^{(n)} P(\theta^{(n)})) P(D | G=k, \m \theta) \nonumber \\
	& = \prod_{n \in \mathcal{T}_k} \frac{\prod_{j \in \textrm{ch}(n)  } \Gamma(N_{j | n} + \alpha^{(n)} ) )}{
		\Gamma(\sum_{j \in \textrm{ch}(n)  } N_{j | n} + |\textrm{ch}(n)|\alpha^{(n)} )	
	}. \label{eq:3}	
\end{align}

\subsection{Priors} \label{sec:priors}
For most choices of concentration parameters $\alpha^{(n)}$, the probability tree model will learn a causal orientation from observational data alone, and therefore create a bias towards a particular causal orientation prior to performing interventions. To prevent this, the concentration parameters should be chosen so that: 
\begin{align}
	\frac{1}{S} = P(G = k | D^\text{observational} ).
\end{align}
To ensure this, suffice it to select the concentration parameter in a node $n$ to be proportional to the number of descendants of $n$ divided by the number of immediate children, i.e. 
\begin{align}
	\alpha^{(n)} = \frac{ |\{\mbox{descendants of $n$}\}| }{|\textrm{ch}(n)|}\alpha \label{eq:priors}
\end{align} 
for a common factor $\alpha > 0$. This is easily seen by noting that most factors in \cref{eq:3} cancel, and the marginal likelihood simply reduces to a product over the leaves:
\begin{align}
	P(D^\text{observational}| G=k) = 
	\frac{ \prod_{n \in \mathcal T_k, \ j \text{ is a leaf} } \Gamma(N_{j | n} + \alpha  ) }{
		\Gamma(N + |\mathcal T| \alpha )	
	}.\nonumber
\end{align}
Thus, the probability of a given causal hypothesis $\mathcal T_k$ computed using \cref{eq:1a} will depend on the data $D$ and a single parameter $\alpha > 0$.

\subsection{Active learning}
The amount of evidence, measured in nats, in favor of a hypothesis $h$ relative to the alternative is $\frac{\log P(h | D)}{\log P(\neg h | D)}$~\cite{jaynes2003probability}. A natural criterion by which to choose between interventions is how much they are expected to change the evidence in favor of the true causal hypothesis. Since the agent only has access to limited data, we distinguish between the agent's estimate of the probability assignment, indicated by the symbol $Q$, and the true probability assignment indicated by $P$.

Suppose the agent performs an intervention $\doo(X_\ell = x_\ell) = \hat x_\ell$ and observes the realization $\mathbf x$ of $\mathbf X$. We define the information gained in favor of a causal hypothesis $G=k$ from observing the effect of $\hat x_\ell$, in the context of an existing data set $D$, as the increase in evidence in favor of $k$:
\begin{align}
	& I(k | D, (\mathbf{ x}, \hat x_\ell ) ) \nonumber \\
	& = \log \frac{Q(k | D, (\mathbf{x}, \hat x_\ell) ) }{\sum_{k' \neq k}^{S} Q(k' | D, (\mathbf{x}, \hat x_\ell) ) } - \log \frac{Q(k | D) }{\sum_{k' \neq g}^{S} Q(k' | D ) }. \label{eq:4}
\end{align}
In the above, the symbol $Q(k|D)$ refers to the agent's belief as computed using \cref{eq:1a} and \cref{eq:3}.  Note that this quantity reflects the agent's internal belief about the truth of a hypothesis $G=k$.

Assume that the (true) causal orientation is among the $S$ causal hypotheses. The chance $k$ is the (true) causal hypothesis given the data is $P( G=k | D,\m \theta)$, where importantly we use $P$ to signify this quantity is computed using the true model of the system (i.e, the likelihood is computed using \cref{eq:5} with the true values of $\theta^{(n)}$).

Thus, the information gain in favor of the \emph{true} causal hypothesis is given as the expected gain in information for a given hypothesis $k$, computed using \cref{eq:4}, weighted by the chance that $k$ is actually the true hypothesis $P(G=k | D, \m \theta)$. Specifically, we define the \emph{actual information gain} as: 
\begin{align}
	\Delta_{\hat x_\ell}^\text{Actual} & = 
	\mathbb{E}_{k | D, \mathbf{x}_\ell | g, D,\hat x_\ell } \left[ I(k | D, (\mathbf{x}, \hat x_\ell) \right] \nonumber \\
	& = \sum_{k=1}^S\! P(k| D,\!\m \theta) \sum_{ \mathbf{x} }\!\! P(\mathbf{x} | k,\! \hat x_\ell) I(k | D,  (\mathbf{ x}, \hat x_\ell ) ). \label{eq:Deltaactual}
\end{align}
This quantity represents the actual information gain experienced by the agent under the model assumptions, and is what we ideally would wish to compute. However, since it depends the true probabilities $P$, we instead define the \emph{expected gain} as:
\begin{align}
	\Delta_{\hat x_\ell }^\text{Expected} & = 
	\sum_{k=1}^S Q(k| D) \sum_{ \mathbf{x} } Q(\mathbf{x} | k, \hat x_\ell, D) I(k | D,  (\mathbf{ x}, \hat x_\ell ) ) \label{eq:expected}
\end{align}
which can be computed using known quantities prior to interventions. The method therefore simply ranks the possible interventions using \cref{eq:expected} and selects the intervention with the highest expected gain.

\subsection{The two-variable case}
It is instructive to consider the simple case of $m=2$ variables $X_1$ and $X_2$, and where the data $D$ only consists of observational data. We consider the case of two causal orientations, $k=1$ corresponding to $X_1 \rightarrow X_2$ and $k=2$ corresponding to $X_2 \rightarrow X_1$, and without loss of generality assume an intervention $\hat x_1$ is performed on $X_1 = x_1$ and we observe $X_2 = x_2$. In this case, it follows from symmetry that $I(k=1|D,(\mathbf{x}, \hat x_\ell)) = -I(k=2|D,(\mathbf{x}, \hat x_\ell))$ and \cref{eq:4} becomes
\begin{align}
	I(k=1| D, (x_2, \hat x_1) ) & = \log \frac{Q(k=1| D, (x_2, \hat x_1) ) }{
		Q(k=2| D, (x_2, \hat x_1)
	} \nonumber  \\ 
	& = 
	\frac{Q(D, (x_2, \hat x_1)| k=1) }{
		Q(D, (x_2, \hat x_1)| k=2)
	}.
\end{align} 
Using \cref{eq:3}, and noting that the single intervention only changes the pseudo-counts $N_{j|n}$ by a single value, the expression simply reduces to the $\alpha$-robust estimate of the probabilities:
\begin{align}
	I(k=1| D, (x_2, \hat x_1) ) & = \log \frac{Q(x_2 | x_1,D) }{ Q(x_2 | D) } \\
	\text{where: }\ Q(x_2 | x_1,D) & = \frac{n(x_1,x_2) + \alpha }{ \sum_{x_2 \in \mathcal{V}_2 } n(x_1,x_2) + |\mathcal{V}_1| \alpha }, \nonumber \\
	Q(x_2 |D) & = \frac{\sum_{x_1 \in \mathcal{V}_1 } n(x_1,x_2) + |\mathcal{V}_1|\alpha }{ N + |\mathcal{V}_1| |\mathcal{V}_2| \alpha }.\nonumber
\end{align} 
In this expression, $n(x_1,x_2)$ is the number of observations in $D$ where $X_1=x_1$ and $X_2=x_2$. 

Similarly, the outer expectations in \cref{eq:Deltaactual} over $\mathbf{x} = (x_1,x_2)$ are easily expressed using the conditional/marginal probabilities, for instance:
\begin{subequations}
	\begin{align}
		P(\m x | k=1, \hat x_1) &  = 	P(X_2 =x_2|X_1 = \hat x_1)\delta_{x_1,\hat x_1},\\
		P(\m x | k=2, \hat x_1) & = 	P(X_2 =x_2)\delta_{x_1,\hat x_1}. 	
	\end{align}
\end{subequations}
The actual expected gain of information in favor of the true causal direction, when an intervention $\hat x_1$ is performed on $X_1 = x_1$, is therefore
\begin{align}
	\Delta^\text{Actual}_{\hat x_1} = \frac{1}{2} \sum_{x_2} \left[P(x_2 | x_1) - P(x_2)   \right]  
	\log \frac{ Q(x_2 | x_1, D) }{ Q(x_2 | D) } \label{eq:Jactual}
\end{align}
The expression for the expected gain is even simpler, and can be expressed using the Jeffrey divergence~\cite{jeffreys1998theory} between the conditional and marginal distributions $D_J(p;q) = \sum_{x} (p(x)-q(x))\log \frac{p(x)}{q(x)}$:
\begin{align}
	\Delta^\text{Expected}_{\hat x_1} = D_J( Q(X_2 | x_1, D); Q(X_2 | D) ). 	\label{eq:9}
\end{align}

The intuition behind this expression is that for two variables, the suitability of an intervention is assessed based on how much the conditional distribution differs from the marginal, as measured by the Jeffrey divergence which corresponds to the case where an intervention may give a particularly surprising result. The Jeffrey divergence may become very large when the probabilities are close to zero, however, this simply corresponds to the case where certain interventions can lead to configurations which are highly atypical for the system, and therefore are highly informative with respect to the causal orientation. 

This does present a potential problem when the exact probability $P$ in \cref{eq:Jactual} is replaced by the finite-sample approximations in \cref{eq:9}. This means that the regularization parameter $\alpha$ can be expected to play an important role. However, in our experiments, we found that values between $\alpha=1$ or $\alpha=2$ worked well.

\begin{figure*}[t!]
	\centering
	\includegraphics[width=.33\linewidth]{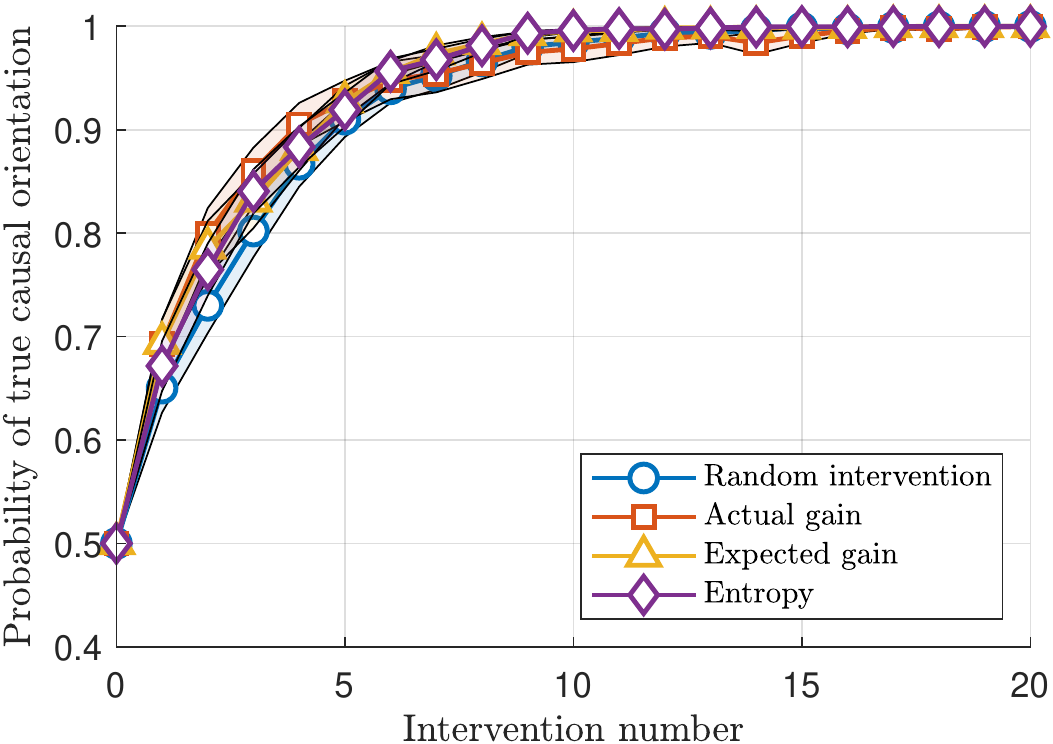}~
	\includegraphics[width=.33\linewidth]{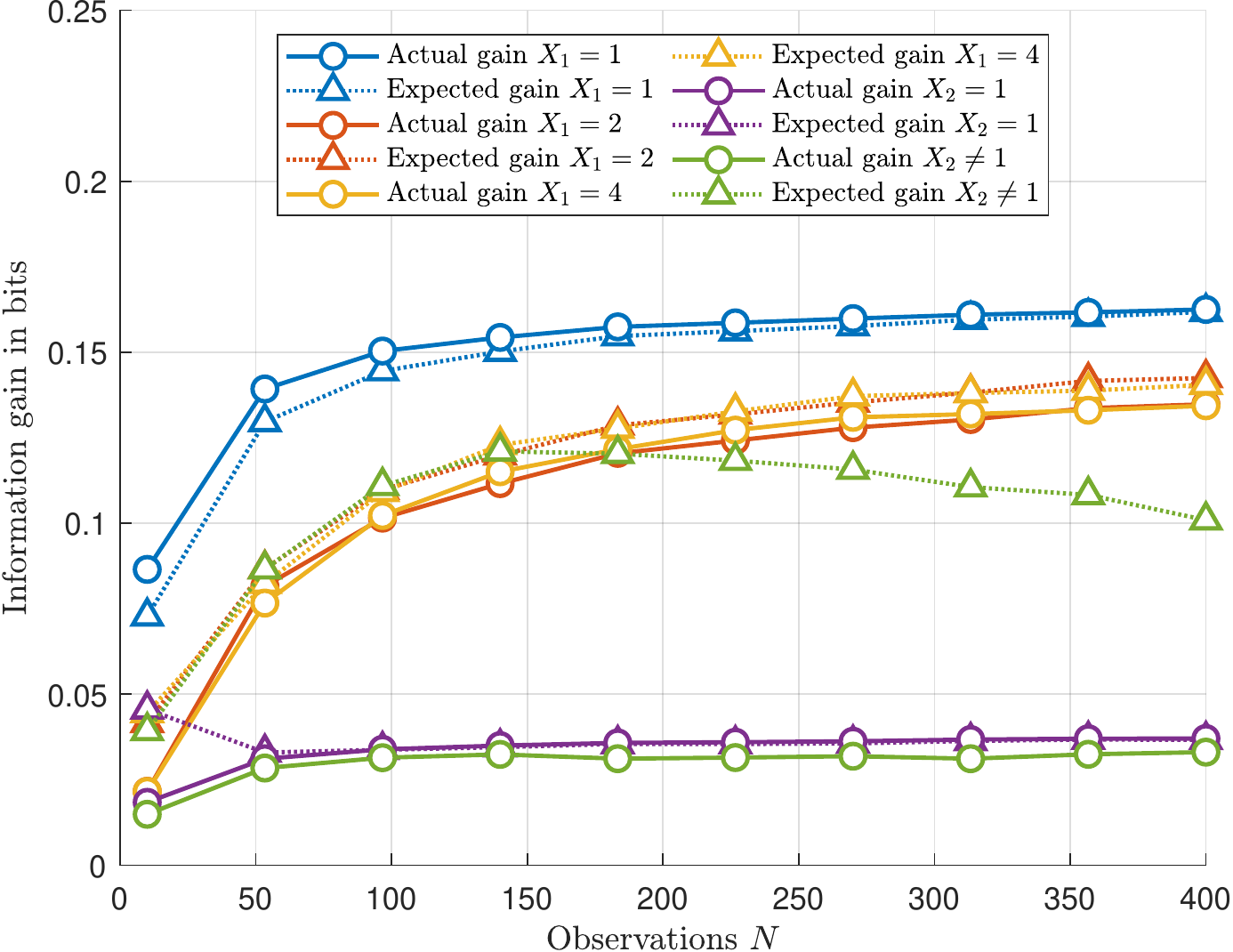}~
	\includegraphics[width=.33\linewidth]{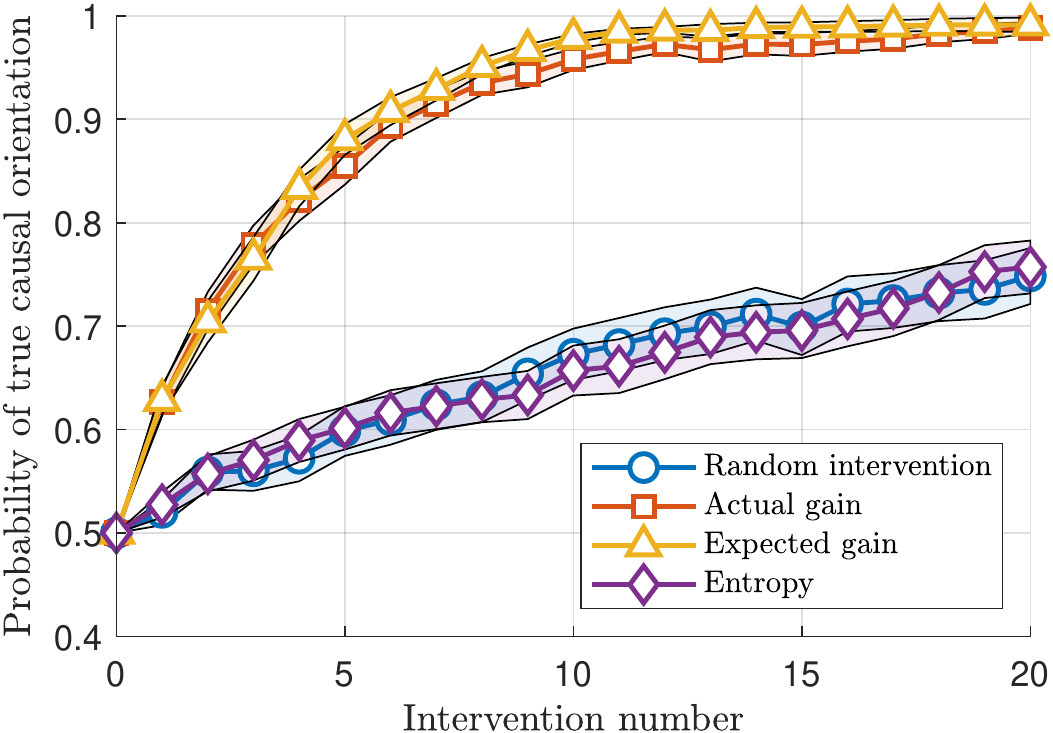}
	\caption{(a) Probability of determining the true causal orientation in the perfectly correlated two-variable problem described in \cref{sec:twovariables}. The methods are evaluated using a fixed number of observations $N$ and varying number of interventions. The different interventions provide nearly the same information gain. (b) Illustration of the expected/actual gain in information for different interventions in the asymmetric two-variable problem. The intervention $X_1=1$ is predicted to offer more information than the other across a varying number of observations $N$. In all cases, the expected gain  \cref{eq:expected} is close to the idealized exact gain \cref{eq:Deltaactual}. The result is born out when using the active-learning method in (c), where the proposed active-learning scheme uses far fewer interventions to determine the true causal orientation. 
	}\label{fig4}
\end{figure*}

\section{Experiments} \label{sec:simulations}
The most relevant work to compare against is the active learning method for causal Bayesian networks proposed by \cite{tong2001active} (\emph{Entropy} in the following). The method uses a cost-function based approach to rank interventions based on their expected outcome, and the cost-function is constructed as follows: For each possible edge $i,j$ in the Bayesian network, there are three possible outcomes ($X_i \rightarrow X_j$, $X_j \rightarrow X_i$ and no edge between $i,j$), which can be considered the outcome of a categorical distribution. Given the current data set, consisting of observations and interventions, the method considers the entropy of these categorical distributions, and constructs the cost-function as the expected value of the sum of all such entropies. The main difference in our approach is that we consider the full structure of the causal network, and instead of the entropy we consider the evidence in favor of a given causal orientation. 

We note that the authors propose a sampling-based approach to generate candidate Bayesian networks, however, we will consider a direct implementation of the cost function as this is feasible in our experiments. The probabilities used in computing the entropies are obtained using the same method (i.e., using $\alpha$-soft estimates) as our method. Note that this method is only applicable to probability trees which can be represented as CBNs.

\subsection{Information gain with two variables}\label{sec:twovariables}
Continuing the example with two variables, consider the case where the variables are perfectly correlated and can take $K$ values $|\mathcal{V}_i| = K$. Assume that the joint distribution is symmetric, $p(x_1,x_2) = \frac{\rho}{K} \delta_{x_1,x_2} + \frac{1-\rho}{ K(K-1)}(1-\delta_{x_1,x_2})$, and that the observational data is proportional to $p$, i.e. $n(x_1,x_2) = N p(x_1,x_2)$, the actual information gain from any intervention $\hat x_1$ will be given by \cref{eq:Deltaactual}. In the case $K=2$ it is:
\begin{align}
	\Delta_{\hat x_\ell } & = \frac{1}{2}
	(\rho-\frac{1}{2} ) \log \frac{ N \rho + 2\alpha }{ N(1-\rho) + 2\alpha }.
\end{align}
Thus, in the case of perfectly correlated variables, the actual gain in information from the \emph{first} intervention will be proportional to $\log N$ and independent of the intervention. We simulated the setup in an active learning setting up to $20$ interventions, where actions were selected using either the actual gain, expected gain or the Entropy method described previously using $K=4$ and $\rho = 0.9$. The simulations were hot-started using $N=300$ non-interventional observations and results were averaged over $T=100$ random restarts. The error bars indicate the standard deviation of the mean. The result can be seen in \cref{fig4} (a). All methods performed well, since the symmetry of the problem makes all possible interventions roughly equally informative. 

A more interesting problem is obtained when the joint distribution is asymmetric. Specifically, we consider the case $K=4$, where the true causal orientation is $X_1\rightarrow X_2$, and where $p(1,x_2) = p(4,1) = \frac{\rho}{5}$, and otherwise $p(x_1,x_2) = \frac{1-\rho}{11}$. Both the actual and expected gain of all possible interventions are shown in \cref{fig4}, here averaged over $T=100$ random restarts, and it is clearly seen that the intervention $X_1 =x_1$ is optimal. It is notable that when probabilities are estimated from relatively few observations, the expected and actual gain still tend to coincide. When the problem is considered in an active learning setting similar to \cref{fig4}, we see that selecting interventions using the expected gain leads to much quicker convergence than random interventions. Somewhat surprisingly, the entropy-based method performs on par with random intervention selection.

\subsection{The Pairs database}
The Pairs database~\cite{mooij2016distinguishing} consists of 108 small data sets intended to provide realistic examples of real-world cause and effect pairs. We limit ourselves to the case $S=2$ and select the variables as the first columns marked as cause and effect. The data set is pre-processed by binning into $K=5$ equiprobable bins, and the results are reported using the weighting procedure (to account for some data sets being similar) as suggested by the authors~\cite{mooij2016distinguishing}. 

The methods for selecting interventions were evaluated by first generating a varying number $N$ of observational data-points, and then for each data set, performing $40$ interventions. In all cases we used $\alpha=1$. The result were averaged over $T=20$ random restarts.

\begin{table}
	\centering
	{\small 
\begin{tabular}{ccccc}
\toprule
 & Actual gain & Expected gain & Entropy & Random \\
\midrule
$N=50$ & \SI{17.19 \pm 0.88}{} & \SI{19.42 \pm 0.9}{} & \SI{27.9 \pm 0.8}{} & \SI{30.63 \pm 0.7}{} \\
$N=100$ & \SI{15.54 \pm 0.83}{} & \SI{16.35 \pm 0.94}{} & \SI{26.35 \pm 0.83}{} & \SI{27.92 \pm 0.75}{} \\
$N=200$ & \SI{14.54 \pm 0.79}{} & \SI{15.14 \pm 0.86}{} & \SI{25.2 \pm 0.91}{} & \SI{25.52 \pm 0.82}{} \\
\bottomrule
\end{tabular}
	}
	\caption{Interventions required in the \emph{Pairs} data set.}\label{tbl:perform}	
\end{table}

As an evaluation metric, we considered the average time until the method was at least 95\% certain about the correct causal orientation (the method defaults to 40 interventions in case the method did not obtain certainty about the true causal orientation). The results can be found in \cref{tbl:perform}. As expected, all methods perform better with more observational data. The overall trend is that the actual gain, computed using \cref{eq:Deltaactual}, outperforms the expected gain \cref{eq:9}. This is to be expected, since the actual gain is more informed about the problem, but in both cases the proposed method learns the causal orientation far quicker than random interventions, or the method of \cite{tong2001active}. 

\subsection{Information gain with three variables}
To test the method in a more challenging setting, which can nevertheless still be represented as a CBN, we consider the case of $m=3$ variables, $X_1$,$X_2$,$X_3$, and where the actual causal relationship is $X_1 \rightarrow X_2,X_3$ and $(X_1,X_2) \rightarrow X_3$. We consider the problem of determining this causal relationships among all graphs obtained by permuting the node labels, i.e. there are $S=6$ possible graphs. This can be represented as a probability tree similar to \cref{fig3svg}, but where the root has six children. 

Each variable could take $|\mathcal V_i| = 6$ values, and to avoid the case where the variables are highly correlated (and all interventions have roughly the same value), we selected the joint distribution by selecting an arbitrary sparsity pattern $\tilde p(x_1,x_2,x_3) \in \{0,1\}$ and normalizing $p(x_1,x_2,x_3) = \frac{\rho}{M}  \tilde p(x_1,x_2,x_3) + \frac{1-\rho}{6^3 - M} (1 - \tilde p(x_1,x_2,x_3))$ where $M = \sum_{x_1,x_2,x_3} \tilde p(x_1,x_2,x_3)$. Details can be found in the supplementary code, and the qualitative outcome is not sensitive to this choice. 

\begin{figure}[t!]
	\centering
	\includegraphics[width=.9\linewidth]{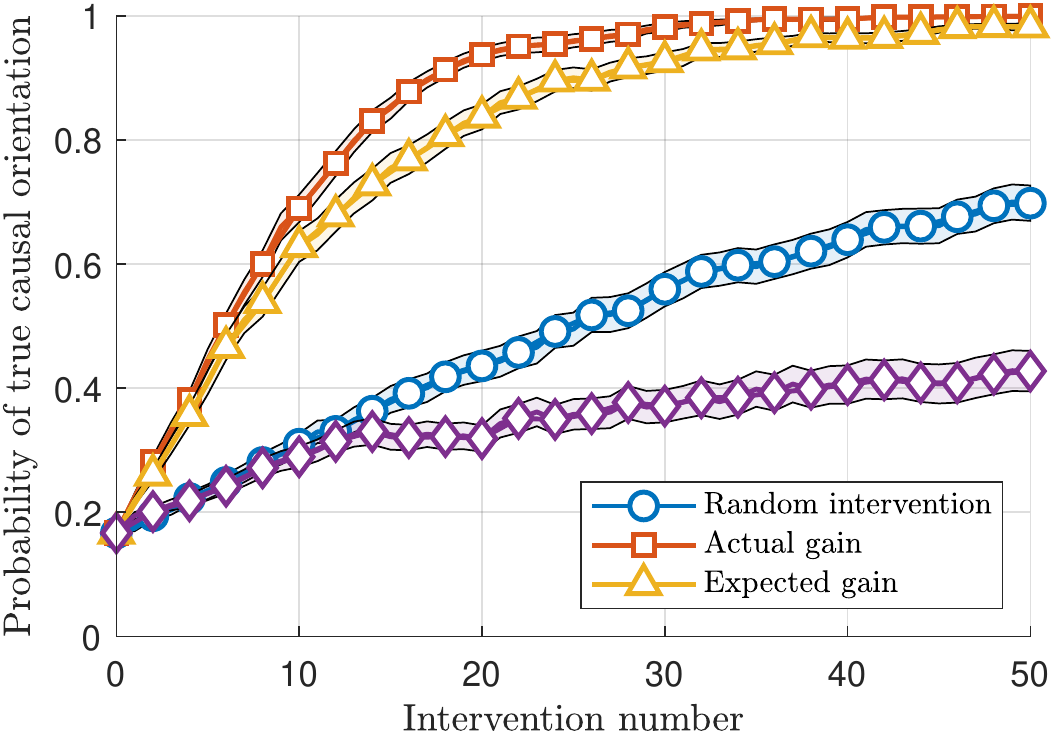}	
	\caption{Result of selecting between the six possible fully-connected CBNs defined on a three-variable problem where each variable can take 6 values using a similar setup as considered in \cref{fig4} (a,b). In this case, our method (expected gain) outperforms random selection, and the entropy-based cost function tends to get stuck on sub-optimal interventions. }\label{fig5}
\end{figure}

Next, we simulated the methods using similar settings as in the two-variable case and using $\rho = 0.9$. The results can be found in \cref{fig5} and $N=300$. Since the probabilities are estimated using relatively less data, the difference between the expected and actual gain is relatively larger. However, selecting interventions according to the expected gain performs far better than random intervention selection. The results are averaged over $T=15$ random restarts. 

Surprisingly, the entropy method performs as well than random intervention selection. We attribute this to our use of a deterministic version of the objective function, and believe that the method would perform better with increasing randomization.

\subsection{Context-dependent causality}
Finally, we consider a context-dependent causal problem which cannot be represented using a CBN. We considered $m=3$ variables each capable of taking $|\mathcal{V}_i| =3$ values. We considered three possible causal hypotheses: In the first, the nodes were ordered as in the previous example, i.e. $X_1 \rightarrow X_2, X_3)$ and $(X_1,X_2) \rightarrow X_3$. In the other two hypotheses, we let $X_1$ determine the causal order of $X_2$ and $X_3$, so that in the first case $X_1=1$ the causal order between $X_2$ and $X_3$ were reversed (similar to the example in \cref{fig2svg2} (d)), and in the other case the causal re-ordering of $X_2$ and $X_3$ occurs when $X_1=2$. The reasoning behind including two context-dependent cases is to ensure that the method does not simply distinguish between context-dependent effect vs. no context-dependent effect. We consider the correlated case, $\tilde p(x_1,x_2,x_3) = \delta_{x_1, x_2}\delta_{x_2,x_3}$ to avoid unintended bias.

\begin{figure}[t!]
	\centering
	\includegraphics[width=.9\linewidth]{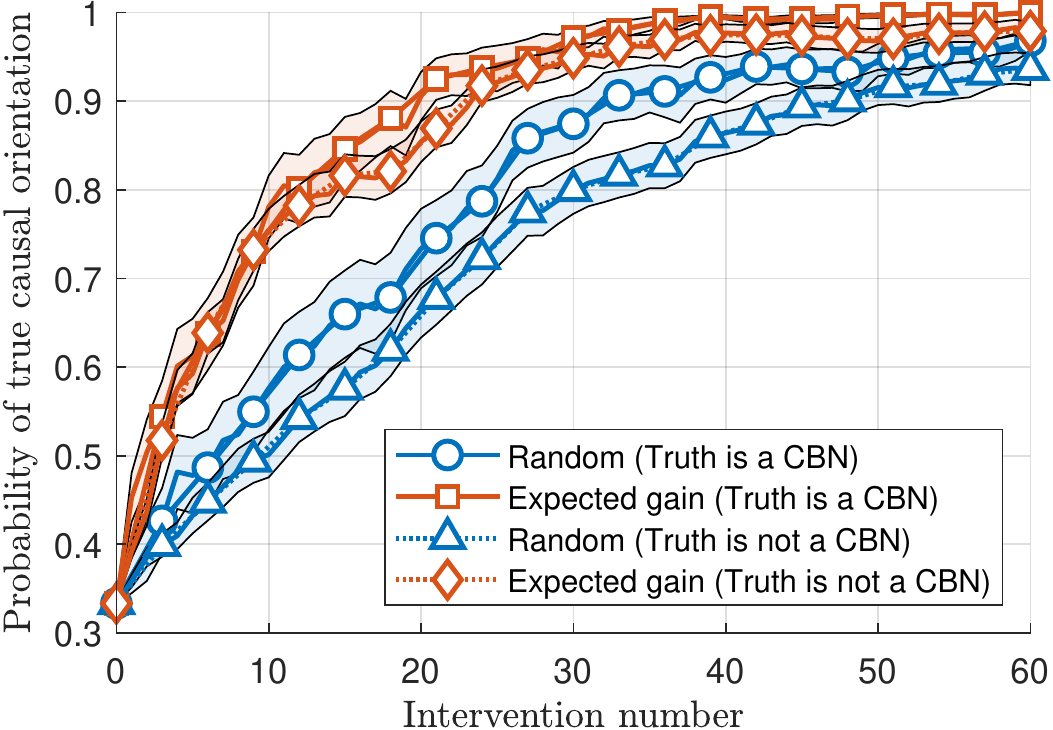}
	\caption{A context-dependent intervention-selection problem defined over three variables. The two conditions correspond to the case where the actual graph is either a CBN or contains a context-dependent causal relation. Our method uses about half as many samples as random intervention selection.  }\label{fig6}
\end{figure}

The methods were run using a similar setup as before, using $N=400$ observations, $\rho=0.9$, and were averaged over $T=400$ random restarts, see \cref{fig6}. We included both the case where the true causal graph had a context-dependent causal effect, and the case where it did not. The performance of the method was similar in the two cases, and shows that true causal orientation can be determined to a given degree of certainty using roughly half as many samples, compared to the case where random interventions are used. We do not include results from the entropy method as it is not applicable to context-dependent effects.

\section{Discussion and conclusion}
Probability trees provide a conceptually simple representation of causal relationships, which is nevertheless strictly more general than SCMs and CBNs by allowing context-dependent causal relationships. This flexibility redefines the problem of causal induction, giving it a self-contained formulation as an inference problem in a single probability tree.

Despite this, probability trees have seen very limited use as a means to represent causal relationships up to this point~\cite{genewein2020algorithms}. A likely reason is that, in contrast to CBNs and SCMs, the independence properties of the variables are not visually apparent from the tree, and although the effect of interventions and counterfactuals can be predicted efficiently from the tree algorithmically, probability trees do not provide convenient tools, such as the do-calculus~\cite{pearl2000models}, which makes them more difficult to use. However, in machine learning there is an increasing interest in learning causal relationships automatically from data~\cite{scholkopf2019causality},  so these drawbacks are less pronounced. This also means that methods for finding the correct tree structure for a given problem can be expected to become a key challenge in furthering the use of probability trees.

In this paper, we have presented the first such method, to our knowledge.  Our approach formulates the problem of causal induction as an inference problem in a single probability tree. We can then define the potential gain in information given the current data using Bayes theorem, and select interventions which are associated with the greatest expected gain. We evaluated the method both by comparing it to a random intervention selection and a deterministic implementation of the cost function from \cite{tong2001active}. In our experiments, the proposed method is able to find the correct causal ordering using fewer samples than the alternative methods, both in the case of relationships that could be represented using CBNs and in context-dependent relationships. 

The computational cost is determined by the joint sum over all causal hypotheses and potential outcomes of an intervention. For larger problems, this cost can quickly become prohibitively large, but this can be overcome by only considering hypotheses which are close to the current most likely hypothesis (as in \cite{tong2001active}), and by replacing the  sum over potential outcomes with a finite sample.

\bibliographystyle{IEEEtran}
\bibliography{IEEEabrv,library2}

\begin{thebibliography}{10}
\providecommand{\url}[1]{#1}
\csname url@samestyle\endcsname
\providecommand{\newblock}{\relax}
\providecommand{\bibinfo}[2]{#2}
\providecommand{\BIBentrySTDinterwordspacing}{\spaceskip=0pt\relax}
\providecommand{\BIBentryALTinterwordstretchfactor}{4}
\providecommand{\BIBentryALTinterwordspacing}{\spaceskip=\fontdimen2\font plus
\BIBentryALTinterwordstretchfactor\fontdimen3\font minus
  \fontdimen4\font\relax}
\providecommand{\BIBforeignlanguage}[2]{{%
\expandafter\ifx\csname l@#1\endcsname\relax
\typeout{** WARNING: IEEEtran.bst: No hyphenation pattern has been}%
\typeout{** loaded for the language `#1'. Using the pattern for}%
\typeout{** the default language instead.}%
\else
\language=\csname l@#1\endcsname
\fi
#2}}
\providecommand{\BIBdecl}{\relax}
\BIBdecl

\bibitem{ijcai2019-199}
\BIBentryALTinterwordspacing
Y.~Wu, L.~Zhang, and X.~Wu, ``Counterfactual fairness: Unidentification, bound
  and algorithm,'' in \emph{Proceedings of the Twenty-Eighth International
  Joint Conference on Artificial Intelligence, {IJCAI-19}}.\hskip 1em plus
  0.5em minus 0.4em\relax International Joint Conferences on Artificial
  Intelligence Organization, 7 2019, pp. 1438--1444. [Online]. Available:
  \url{https://doi.org/10.24963/ijcai.2019/199}
\BIBentrySTDinterwordspacing

\bibitem{chiappa2019path}
S.~Chiappa, ``Path-specific counterfactual fairness,'' in \emph{Proceedings of
  the AAAI Conference on Artificial Intelligence}, vol.~33, no.~01, 2019, pp.
  7801--7808.

\bibitem{everitt2021agent}
T.~Everitt, R.~Carey, E.~Langlois, P.~A. Ortega, and S.~Legg, ``Agent
  incentives: A causal perspective,'' in \emph{Proceedings of the Thirty-Fifth
  AAAI Conference on Artificial Intelligence,(AAAI-21). Virtual. Forthcoming},
  2021.

\bibitem{richens2020improving}
J.~G. Richens, C.~M. Lee, and S.~Johri, ``Improving the accuracy of medical
  diagnosis with causal machine learning,'' \emph{Nature communications},
  vol.~11, no.~1, pp. 1--9, 2020.

\bibitem{prosperi2020causal}
M.~Prosperi, Y.~Guo, M.~Sperrin, J.~S. Koopman, J.~S. Min, X.~He, S.~Rich,
  M.~Wang, I.~E. Buchan, and J.~Bian, ``Causal inference and counterfactual
  prediction in machine learning for actionable healthcare,'' \emph{Nature
  Machine Intelligence}, vol.~2, no.~7, pp. 369--375, 2020.

\bibitem{bareinboim2015bandits}
E.~Bareinboim, A.~Forney, and J.~Pearl, ``Bandits with unobserved confounders:
  A causal approach,'' \emph{Advances in Neural Information Processing
  Systems}, vol.~28, pp. 1342--1350, 2015.

\bibitem{dasgupta2019causal}
I.~Dasgupta, J.~Wang, S.~Chiappa, J.~Mitrovic, P.~Ortega, D.~Raposo, E.~Hughes,
  P.~Battaglia, M.~Botvinick, and Z.~Kurth-Nelson, ``Causal reasoning from
  meta-reinforcement learning,'' \emph{arXiv preprint arXiv:1901.08162}, 2019.

\bibitem{de2019causal}
P.~de~Haan, D.~Jayaraman, and S.~Levine, ``Causal confusion in imitation
  learning,'' \emph{Advances in Neural Information Processing Systems},
  vol.~32, pp. 11\,698--11\,709, 2019.

\bibitem{yue2020interventional}
Z.~Yue, H.~Zhang, Q.~Sun, and X.-S. Hua, ``Interventional few-shot learning,''
  \emph{Advances in Neural Information Processing Systems}, vol.~33, 2020.

\bibitem{pearl2018book}
J.~Pearl and D.~Mackenzie, \emph{The book of why: the new science of cause and
  effect}.\hskip 1em plus 0.5em minus 0.4em\relax Basic books, 2018.

\bibitem{scholkopf2019causality}
B.~Sch{\"o}lkopf, ``Causality for machine learning,'' \emph{arXiv preprint
  arXiv:1911.10500}, 2019.

\bibitem{pearl2000models}
J.~Pearl \emph{et~al.}, ``Models, reasoning and inference,'' \emph{Cambridge,
  UK: CambridgeUniversityPress}, vol.~19, 2000.

\bibitem{spirtes2000causation}
P.~Spirtes, C.~N. Glymour, R.~Scheines, and D.~Heckerman, \emph{Causation,
  prediction, and search}.\hskip 1em plus 0.5em minus 0.4em\relax MIT press,
  2000.

\bibitem{pmlr-v6-dawid10a}
\BIBentryALTinterwordspacing
A.~P. Dawid, ``Beware of the dag!'' in \emph{Proceedings of Workshop on
  Causality: Objectives and Assessment at NIPS 2008}, ser. Proceedings of
  Machine Learning Research, I.~Guyon, D.~Janzing, and B.~Schölkopf, Eds.,
  vol.~6.\hskip 1em plus 0.5em minus 0.4em\relax Whistler, Canada: PMLR, 12 Dec
  2010, pp. 59--86. [Online]. Available:
  \url{https://proceedings.mlr.press/v6/dawid10a.html}
\BIBentrySTDinterwordspacing

\bibitem{shafer1996art}
\BIBentryALTinterwordspacing
G.~Shafer, \emph{The Art of Causal Conjecture}, ser. Artificial
  Management.\hskip 1em plus 0.5em minus 0.4em\relax MIT Press, 1996. [Online].
  Available: \url{https://books.google.dk/books?id=sY7os7OCykUC}
\BIBentrySTDinterwordspacing

\bibitem{genewein2020algorithms}
T.~Genewein, T.~McGrath, G.~Del{\'e}tang, V.~Mikulik, M.~Martic, S.~Legg, and
  P.~A. Ortega, ``Algorithms for causal reasoning in probability trees,''
  \emph{arXiv preprint arXiv:2010.12237}, 2020.

\bibitem{wieczorek2019information}
A.~Wieczorek and V.~Roth, ``Information theoretic causal effect
  quantification,'' \emph{Entropy}, vol.~21, no.~10, p. 975, 2019.

\bibitem{compton2021entropic}
S.~Compton, M.~Kocaoglu, K.~Greenewald, and D.~Katz, ``Entropic causal
  inference: Identifiability and finite sample results,'' \emph{arXiv preprint
  arXiv:2101.03501}, 2021.

\bibitem{janzing2012information}
D.~Janzing, J.~Mooij, K.~Zhang, J.~Lemeire, J.~Zscheischler, P.~Daniu{\v{s}}is,
  B.~Steudel, and B.~Sch{\"o}lkopf, ``Information-geometric approach to
  inferring causal directions,'' \emph{Artificial Intelligence}, vol. 182, pp.
  1--31, 2012.

\bibitem{tong2001active}
S.~Tong and D.~Koller, ``Active learning for structure in bayesian networks,''
  in \emph{International joint conference on artificial intelligence}, vol.~17,
  no.~1.\hskip 1em plus 0.5em minus 0.4em\relax Citeseer, 2001, pp. 863--869.

\bibitem{vajda2014probabilistic}
S.~Vajda, \emph{Probabilistic programming}.\hskip 1em plus 0.5em minus
  0.4em\relax Academic Press, 2014.

\bibitem{winn2012causality}
J.~Winn, ``Causality with gates,'' in \emph{Artificial Intelligence and
  Statistics}.\hskip 1em plus 0.5em minus 0.4em\relax PMLR, 2012, pp.
  1314--1322.

\bibitem{jaynes2003probability}
E.~T. Jaynes, \emph{Probability theory: The logic of science}.\hskip 1em plus
  0.5em minus 0.4em\relax Cambridge university press, 2003.

\bibitem{jeffreys1998theory}
H.~Jeffreys, \emph{The theory of probability}.\hskip 1em plus 0.5em minus
  0.4em\relax OUP Oxford, 1998.

\bibitem{mooij2016distinguishing}
J.~M. Mooij, J.~Peters, D.~Janzing, J.~Zscheischler, and B.~Sch{\"o}lkopf,
  ``Distinguishing cause from effect using observational data: methods and
  benchmarks,'' \emph{The Journal of Machine Learning Research}, vol.~17,
  no.~1, pp. 1103--1204, 2016.

\end{thebibliography}

\end{document}